\documentclass[letterpaper, 10pt, conference]{ieeeconf}  %

\IEEEoverridecommandlockouts                              %
\usepackage{graphics} %
\usepackage{epsfig} %
\usepackage{amsmath} %
\usepackage{amssymb}  %

\title{\LARGE \bf
Crowd Counting via  Weighted VLAD on Dense Attribute Feature Maps
}

\author{Biyun Sheng$^{1}$, Chunhua Shen$^{2}$, Guosheng Lin$^{2}$, Jun Li$^{1}$, Wankou Yang$^{1}$, Changyin Sun$^{1}$%
\thanks{*This work was done when B. Sheng was visiting The University of Adelaide.}%
\thanks{$^{1}$ School of Automation, Southeast University, Nanjing 210096, China.}%
\thanks{$^{2}$ School of Computer Science, The University of Adelaide, SA 5005, Australia.
       }%
}

\begin{document}

\maketitle

\begin{abstract}

  Crowd counting is an important task in computer vision, which has many applications in    video surveillance.
  Although the regression-based framework has achieved great improvements for crowd counting, how to improve the discriminative power of image representation is still an open problem.  Conventional holistic features used in crowd counting often
  fail to capture semantic attributes and spatial cues of the image. In this paper, we propose integrating semantic information into learning locality-aware feature sets for accurate crowd counting. First, with the help of convolutional neural network (CNN), the original pixel space is mapped onto a dense attribute feature map, where each dimension of the pixel-wise feature indicates the probabilistic strength of a certain semantic class. Then, locality-aware features (LAF) built on the idea of spatial pyramids on neighboring patches are proposed to explore more spatial context and local information. Finally, the traditional VLAD encoding method is extended to a more generalized form in which diverse coefficient weights are taken into consideration. Experimental results validate the effectiveness of our presented method.

\end{abstract}

\tableofcontents

\clearpage

\section{Introduction}

Crowd counting, which is defined as the number of people in a crowd, has recently drawn great attention due to its wide application in the field of video surveillance, particularly in  public security \cite{c1,c2}. Among exsiting approaches \cite{c3,c5,c6,c7,c8,c9}, the regression-based framework that learns a regression function  between image representations and count numbers has gained considerable interest \cite{c9}. The majority of these works usually construct crowd models from patterns of holistic hand-crafted features  \cite{c1,c9,c10} such as foreground segment features, internal edge features and texture features.

Despite the success of hand-crafted features for crowd counting, there still exists two weaknesses. Firstly, they simply characterize the visual contents without capturing the underlying semantic information, which may lead to unsatisfactory
performance.
In the literature, there is limited work on exploring attribute features for crowd counting.  Therefore,
it is worth  exploring the problem how to improve the discriminative power of feature representations by sufficiently mining the rich
semantic information. Secondly, the holistic features directly obtained from entire images are
typically unable to capture locally spatial information and describe the diversity in the
crowd distribution, density and behaviors. Thus, it is important to encode spatial cues into the feature learning process for crowd counting.

In this work, we attempt to design a novel image representation which takes into consideration semantic attributes as well as spatial cues.
In order to characterize the distribution of people number, we define semantic attributes at the pixel level and learn the semantic feature map
via deep convolutional neural network (CNN). Then, a high-level concept named locality-aware feature (LAF) in the abstract semantic attribute feature map is presented to describe the spatial information in crowd scenes. Fig.~1 shows pixel-level attribute feature maps corresponding to two frames of the UCSD dataset. Every point in the feature map can be expressed as a vector of attributes, each element of which is a certain semantic class probability. After obtaining local descriptors LAF from adjacent sampled cells over the semantic feature map, our pipeline adopts an improved method Weighted VLAD (W-VLAD) to encode  features into image representations.

 \begin{figure}[t!]
      \centering       \includegraphics[scale=0.475]{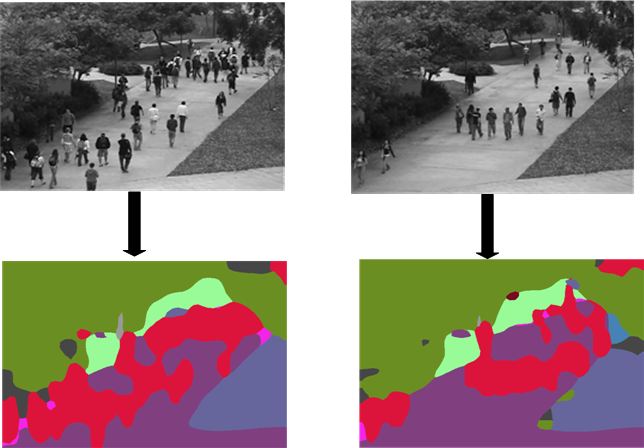}
      \caption{The illustration of two video frames sampled from the UCSD dataset (the first row) and  the corresponding probability maps over semantic feature space
      (the second row). The scene contains some representative semantic attributes (\textit{e.g.},
    person, road,  trees, etc.) shown in different colors.}
      \label{figurelabel}
   \end{figure}

Compared with the traditional method that directly obtains the image representation by aggregating holistic features \cite{c1,c9,c10}, our pipeline that combines LAF with W-VLAD is more discrimative for creating instance representations.
It can not only adapt to complex diversity in crowd \cite{c1}, but also retain more useful information. The comparison
between our proposed method and the traditional ones is shown in Fig.~2.

\begin{figure*}[t!]
      \centering
      \includegraphics[width=0.75\textwidth]{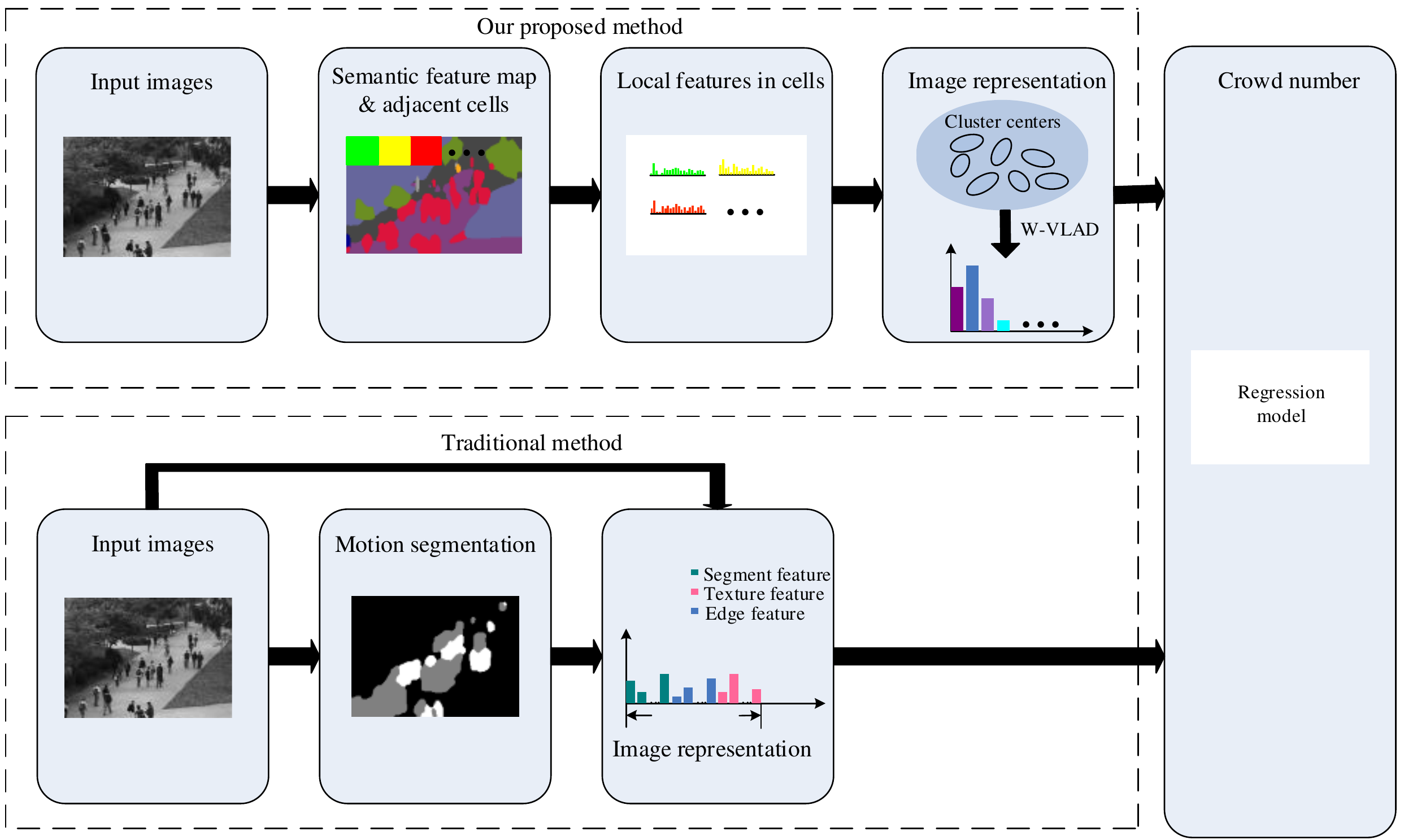}
      \caption{Comparison between our approach and the conventional methods
      \cite{c1,c9,c10}. The first row shows our proposed process including semantic feature map construction, local features learning and W-VLAD encoding, while the second row is the existing work with simple concatenation of coarse motion segmentation feature, edge feature and texture feature directly from the entire image.}
      \label{figurelabel}
\end{figure*}

In summary, the main contributions of this paper are as follows.
\begin{itemize}
  \item
  For the first time, a pixel-level semantic feature map learned by a deep learning model is constructed for local features extraction.

\item
 A novel descriptor named LAF is proposed to combine both advantages of explicit semantic incorporation and spatial context encoding.

\item
 An improved VLAD scheme termed W-VLAD is applied for encoding descriptors into the  final representation.
\item
 Extensive experiments on benchmarks demonstrate the effectiveness of our method, even when people in the frames are hard to identify.
\end{itemize}

The rest of this paper is organized as follows. After reviewing the related work in Section
\uppercase\expandafter{\romannumeral2}, we introduce semantic locality-aware   features in Section \uppercase\expandafter{\romannumeral3}. Then,  the thorough experimental evaluations are carried out in Section
\uppercase\expandafter{\romannumeral4}. Finally, the paper is concluded in Section
\uppercase\expandafter{\romannumeral5}.

\section{Related Work}

\textbf{Counting methods.} Most previous work on crowd counting can be divided into three categories: counting pedestrians by detection \cite{c17,c18,c19}, clustering \cite{c3,c5} and regression \cite{c1,c6,c9}. The performance of detection approaches highly depend on the whole pedestrian result which is suppressed in densely crowded scenes with significant occlusion. Clustering-based methods that coherent motion patterns are clustered to estimate the crowd number only work well for video frames at a high rate. Considering severe occlusions between people in the crowd scene and the requirements of high efficiency for real applications,  many researchers utilize regressors trained on low-level features to predict the global counts \cite{c1,c6,c7,c8,c9}. Most of these methods are based on low-level features proposed in \cite{c9}. With the ever-growing data volume and computational capability, deep learning has begun a useful tool in several surveillance applications such as crowd behavior analysis \cite{c21} and crowd segmentation \cite{c25}. Zhang \textit{et al.} \cite{c2} firstly try to develop a deep model for  predicting crowd number and the superior performance validates that deep feature is more discriminative than shallow hand-craft one.

\textbf{Attribute learning.} Attributes are mid-level semantic properties of objects which bridge the gap between low-level features and high-level representations \cite{c35}. In recent years, attribute-based representations that describe a target by multiple attribute classes have been widely used to represent objects \cite{c27}, faces \cite{c28} and actions \cite{c29}. The common characteristics shared by different scenes can express more information \cite{c22}.  Shao \textit{et al.} \cite{c21} utilize three attributes "who", "where" and "why" to describe a crowd scene and successfully apply them to analyze crowd behavior. Wang \textit{et al.} \cite{c46} propose that attributes and interdependencies among them can help to improve object recognition performance. Liu \textit{et al.} \cite{c47} present an action recognition framework built upon a latent SVM formulation where latent variables capture the degree of each attribute to its action class. Similarly, a binary cumulative attribute is converted from the count number \cite{c14}, which has realized the crowd counting task effectively. However, cumulative attribute, only a coarse description of the crowd distribution, is not sufficient for the natural scene.

\textbf{Spatial cues encoding.} Recent works demonstrate that it is vital to describe local features with spatial awareness for minimizing the information loss and improving the discriminative power of representations. Usually holistic representations are sensitive to changes in the external environment \cite{c9,c20}. Ryan \textit{et al}. \cite{c1} propose a localized approach that estimates the crowd number within different groups and then obtains the total number by summing up all the group results. Chen \textit{et al.} \cite{c10} concatenate features in all sub-regions of an image and  propose a multi-output regression model to learn the local count. He \textit{et al.} \cite{c48} advocate a spatial pyramid pooling (SPP) layer on the entire convolutional feature map and then generates a fixed-length representation for the following fully-connected layers. With spatial information considered, these methods either require manually local counting or lead to representations of huge length.

\textbf{VLAD encoder.} As a simplified version of FV \cite{c33}, VLAD \cite{c34} only aggregates the first order information  which are displacements between features and the corresponding dictionary element. Because of the low computation cost and superior performance, VLAD has recently been applied in many applications. Ng \textit{et al.} \cite{c52} extract convolutional
features from different layers of CNN and apply VLAD to encode features into a single vector
for image retrieval. Multi-VLAD \cite{c51} aims at constructing and matching VLAD vectors of multi-level images. Peng \textit{et al.} \cite{c50} improves VLAD by optimizing the dictionary and considering high-order statistics.

\section{Locality-aware features in the dense attribute feature map}
The key difference between our approach and previous works is that the input to a regression model is a novel and discriminative representation instead of explicit or  implicit usage of the low-level feature \cite{c1,c6,c9,c10}. As shown in the top part of Fig. 2, our proposed approach consists of feature map construction, local feature extraction and encoding. In this section, we introduce the three aspects in detail.

\subsection{Pixel-level attribute feature map}
We attempt to learn image attributes at the pixel level so that semantic information and the diversity of crowd number are sufficiently described. To the best of our knowledge, this is the first work that extracts frame features based on pixel-level attributes of a crowd scene.

Inspired by the tremendous success of deep convolutional neural network (CNN) in pixel labeling \cite{c11,c12,c13}, we learn the pixel-wise semantic feature map by an off-the-shelf deeply learned CNN model. In this paper, the state-of-the-art  deep semantic segmentation method \cite{c13} is applied to train a model on the cityscape dataset \cite{c49}, in which each pixel is annotated with labels such as \textit{person}, \textit{road}, \textit{tree} and so on. Developed as a deep hierarchical structure, the learned CNN model processes the raw image pixels with several consecutive convolutional layers and contextual deep CRF, the output of which is up-sampled as high-level semantic feature map to model the distribution of people in crowd counting.

Specifically, the deeply learned model is utilized as a mapping model $ \mathcal{F}: \mathcal{X} \rightarrow \mathcal{D} $, where $\mathcal{X} \in  \mathbb{R}^{m \times n}$ is the original image in a pixel space, $ \mathcal{D} \in  \mathbb{R}^{m \times n \times p}$ is the corresponding semantic feature map; notions \textit{m}, \textit{n}, \textit{p} respectively denote the image width, height and the number of defined attribute types. The pixel-wise \textit{p} dimensional vector is the probability of \textit{p} attribute classes.

Instead of low-level simple intensity or texture information, the deeply learned semantic feature map characterizes each pixel with abstract attributes. As shown in Fig. 3, the attribute information indicates the co-occurrence of visual patterns (\textit{e.g., a person is likely to co-occur with the road}), which can provide abundant information for sufficiently describing the crowd scene.

 \begin{figure*}[h!]
      \centering
       \includegraphics[scale=0.42]{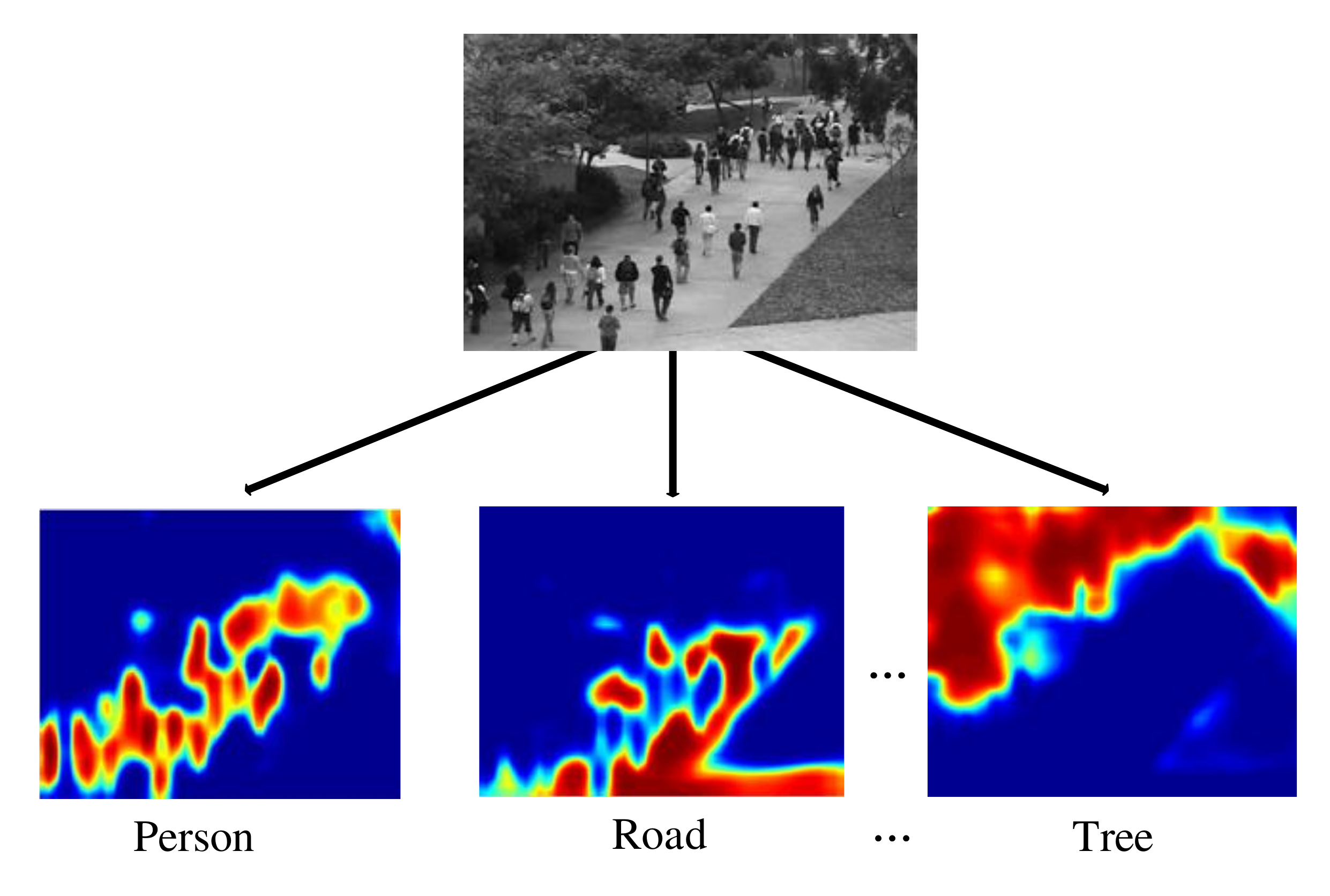}
      \caption{Illustration of pixel-level semantic information corresponding to different attribute classes.}
      \label{figurelabel}
   \end{figure*}

\subsection{Locality-aware features}
On one hand, it is computationally inefficient to take vectors of every pixel as features. On the other hand, computing a feature vector over the entire pixel-wise probability map may lead to large information loss. The most common method is region division as He \textit{et al.} \cite{c48} do, by which the image is partitioned into multiple cells and the concatenation of all cell-based pooling results is taken as the feature. Theoretically the finer the region is partitioned, the more information is retained and the longer the final representation is.

In order to obtain a balance between accuracy and computational complexity, locality-aware features are proposed in this paper. The process of LAF extraction is shown in the right part of Fig. 4 and the procedure is mainly summarized as the following three steps:

1) Given a semantic segmentation feature map  $ \mathcal{D} \in  \mathbb{R}^{m \times n \times p}$,  we first partition it into $N$ cells from which local features are extracted.

2) Each local feature vector of LAF is extracted from each cell, in which spatial pyramid is applied to capture spatial relationship.

The spatial pyramid result of each cell is obtained by concatenating and normalizing the $M$ mean pooling vectors where $M$ is the partition number of a cell. If $p$ is defined as the number of attribute classes, then the local feature in each cell is expressed as $x \in  \mathbb{R}^{d}$, where $d = M p$.
The feature vector $x$ is able to capture the co-occurrence of different semantic attributes.

\begin{figure*}[thpb]
      \centering
       \includegraphics[scale=0.45]{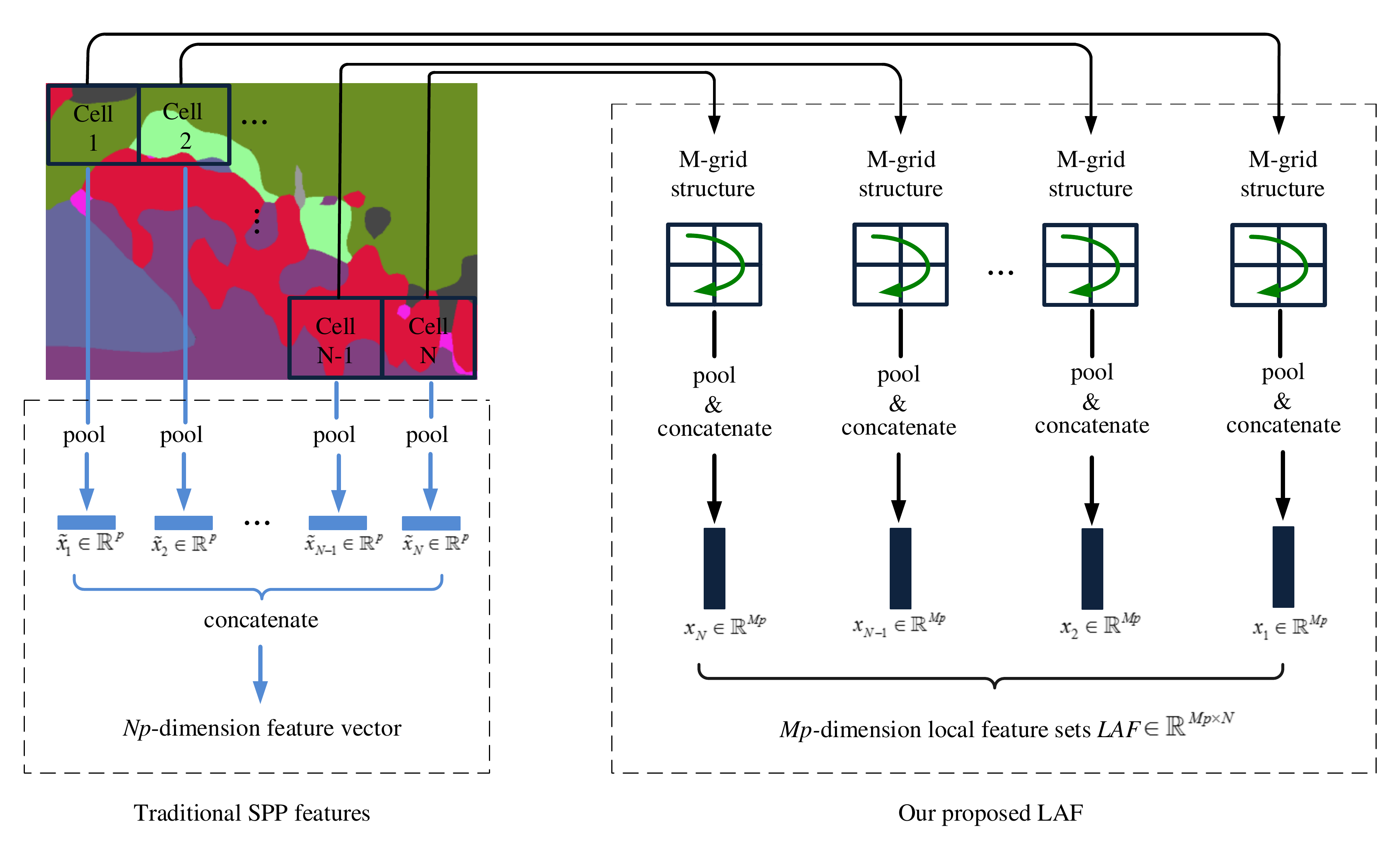}
      \caption{Comparison between LAF extraction and the traditional SPP features. The right part shows LAF extraction in which mean pooling is adopted and the concatenation is operated along the green arrow direction on the M-grid structure (here $M=2 \times 2$); the left part illustrates the traditional SPP features \cite{c12} where spatial information is constructed at the coarse image level.}
      \label{figurelabel}
   \end{figure*}

3) Finally feature vectors of all cells are aggregated as LAF, which can be denoted as:
\begin{equation}
X = [x_{1},x_{2},\cdots,x_{\emph{i}},\cdots,x_{\emph{N}}]\in  \mathbb{R}^{d \times \emph{N}},     1\leq \emph{i} \leq \emph{N} \end{equation}
where ${N}$ is the number of LAF (namely the number of partitioned cells) in the image and $x_{\emph{i}}$ is the feature vector of the \textit{i-th} cell.

In contrast to the SPP feature extraction \cite{c48} which only incorporates the coarse spatial cues at the image level, the set of all $N$ cell-based features are simultaneously and equally used for subsequent feature encoding; spatial pyramid used in LAF is able to maximize the use of local spatial information and characterize context clues. In a word, LAF have advantages of capturing finer location information by localized spatial pyramid and controlling the vector length by aggregating local features.

\subsection{Feature encoder: W-VLAD}
The encoder is utilized to encode a set of low-dimensional local descriptors (LAF here) to a single compact vector. Given a video frame, we first extract LAF $X = [x_{1},x_{2},\cdots,x_{\emph{N}}]\in  \mathbb{R}^{d \times \emph{N}}$ described above. Let $\phi=[\tilde{x}_{1}, \tilde{x}_{2}, \cdots, \tilde{x}_{N}]$ be projected features related to $X$ by PCA+Whitening for better performance \cite{c44} and $\tilde{B}=[\tilde{b}_{1}, \tilde{b}_{2}, \cdots, \tilde{b}_{K}]\in R^{d\times K}$ is the corresponding dictionary learned offline by k-means. The original version of VLAD is expressed as follows:

\begin{equation}
v_{k}=\sum_{i:NN(\tilde{x}_{i})=\tilde{b}_{k}}(\tilde{x}_{i}-\tilde{b}_{k})
\end{equation}

The final representation is gained by concatenating the all ${v}_{k}(k = 1, \cdots ,K)$ vectors with normalization followed. It can be easily seen from the formula (2) that the weights for all residuals are assigned binary values 0 or 1. We try to modify the original VLAD into W-VLAD.

The formula (2) is extended to an alternative form, which is shown in (3):
\begin{equation}
{v}_{k}=\sum_{i=1}^{N}\alpha_{i}^{k}({x}_{i}-{b}_{k})
\end{equation}

For the standard VLAD, the assignment coefficients $\alpha_{i}^{k}$ can be defined as follows:

\begin{eqnarray}&&
\alpha_{i}^{k}=\left\{\begin{array}{cc}
1   & NN({x}_{i})={b}_{k}\\
0   & {\rm otherwise}              \\
\end{array} \right.
\end {eqnarray}

On one hand, the assignment method in (4) is similar to Hard-assignment coding \cite{c39}, which may result in relatively high reconstruction error while coding. On the other hand, the coefficient $\alpha_{i}^{k}$ here can be interpreted as the degree of local feature ${x}_{i}$ to cluster center ${b}_{k}$. In fact, different features may have different relevance to the clusters and it is unreasonable to apply the same weight coefficients even though they are assigned to the same cluster word with the same residual value.

In this paper, we utilize local soft assignment coding (LSAC) \cite{c39} to compute weights for effectiveness and simplicity:

\begin{equation}\label{eq2}
    \alpha_{i}^{k}= \frac{\exp(-\beta d(\tilde{x}_{i},\tilde{b}_{k}))}{\sum_{j=1}^{K} \exp(-\beta d(\tilde{x}_{i},\tilde{b}_{j}))}\\
\end{equation}

\begin{eqnarray}&&
d(\tilde{x}_{i},\tilde{b}_{j})=\left\{\begin{array}{cc}
\|\tilde{x}_{i}-\tilde{b}_{k}\|^{2}   & if \ \tilde{b}_{k}\in \aleph_{\kappa}(\tilde{x}_{i})       \\
\infty       & {\rm otherwise}       \\
\end{array} \right.
\end {eqnarray}
where $\aleph_{\kappa}$ denotes the $\kappa$ nearest neighbors of $\tilde{x}_{i}$ in the dictionary and $\beta$ is the smoothing factor controlling the softness of the assignment.

\section{Experiments}

\noindent \textbf{Datesets.} In order to explore whether our method is effective for the dataset in which persons are not easy to identify, we
introduce a new dataset Caltech 10X (Caltech) \cite{c4} (originally utilized for pedestrian detection) besides the shopping mall dataset (Mall) \cite{c10,c14} and the established UCSD pedestrian (UCSD) \cite{c40}. As shown in Fig. 5, different from Mall and UCSD dataset, persons in some frames from Caltech are not clearly visible. We evaluate the proposed approach on the three datasets in this study for comparative evaluation.

 \begin{figure}[!htb]
      \centering
       \includegraphics[scale=0.45]{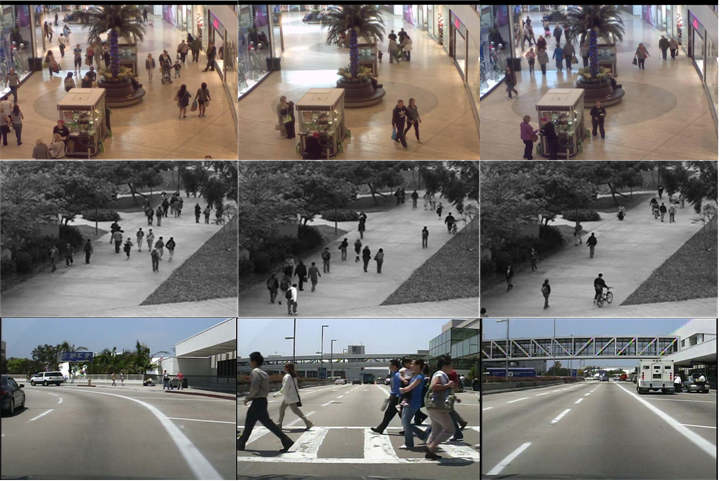}
      \caption{Example frames from three datasets: Row 1, Row 2 and Row 3 are sampled respectively from the
Mall dataset, the UCSD dataset and the Caltech dataset.}
      \label{figurelabel}
   \end{figure}

Table \uppercase\expandafter{\romannumeral1} describes the details of all mentioned datasets. $N_{f}$ means the frame number, D shows the range of crowd number in a frame ROI, and $T_{P}$ indicates the total number of labeled pedestrians.

\begin{table}[h]
\caption{The details of three datasets. }
\label{table_example}
\begin{center}
\begin{tabular}{|c|c|c|c|c|c|c|c|}
\hline
Dataset & $N_{f}$  & {D}  & $T_{P}$\\
\hline
Mall & 2000   & 13-53  & 62325\\
\hline
UCSD & 2000   & 11-46  & 49885\\
\hline
Caltech & 2000 & 6-14  & 15043\\
\hline
\end{tabular}
\end{center}
\end{table}

\noindent \textbf{Evaluation metric.} We use two metrics to evaluate our method, namely mean absolute error (MAE) and mean squared error (MSE). \textit{For both metrics, the lower the values are, the better the experimental performance is}.
\begin{equation}
\begin{aligned}
MAE=\sum_{i=1}^{N_{f}}|y_{i}-\hat{y_{i}}|\\
MSE=\sum_{i=1}^{N_{f}}(y_{i}-\hat{y_{i}})^{2}
\end{aligned}
\end{equation}

\noindent where ${N}_{f}$ is the total number of frames, ${y}_{i}$ is the estimated crowd number in the \textit{ith} frame and $\hat{y_{i}}$ is the corresponding ground truth.

\noindent \textbf{Implementation details.} In Table \uppercase\expandafter{\romannumeral2}, parameters of the second \& third column respectively show a partition number over the entire frame \& the partitioned cell (\textit{B} part of Sec \uppercase\expandafter{\romannumeral3}). The fourth and the last column of Table \uppercase\expandafter{\romannumeral2} empirically defines the dictionary size and the number of nearest neighbors for the W-VLAD encoding process (\textit{C} part of Sec \uppercase\expandafter{\romannumeral3}).

\begin{table}[h]
\caption{The parameters settings for three datasets. }
\label{table_example}
\begin{center}
\begin{tabular}{|c|c|c|c|c|c|c|}
\hline
Dataset & $N$  & $M$ & $K$ & $\kappa$\\
\hline
Mall & 20$ \times $20   & 2$ \times $2  & 100 & 10\\
\hline
UCSD & 20$ \times $20   & 2$ \times $2 & 100 & 10\\
\hline
Caltech & 20$ \times $20 & 2$ \times $2 & 80 & 10\\
\hline
\end{tabular}
\end{center}
\end{table}

\noindent \textbf{Comparative study.} Under the regression-based framework, we try to explore the performance of different image representations. In order to investigate the effectiveness of all parts in our proposed method, we conduct a series of baselines:
\begin{itemize}
\item Holistic Feature (HF). The image representation is obtained by direct mean pooling over the entire dense attribute feature map.
\item SPP Feature (SPPF). The image representation is similar to HF except applying spatial pyramid on the entire feature map.
\item LAF + VLAD (LFV). The image representation is gained by using the original VLAD method to encode our proposed locality-aware features LAF.
\end{itemize}

\subsection{Mall dataset}

Captured at different time of a day, the Mall dataset is challenging for its diverse crowd densities (from spare to dense) and various activity patterns (from static to moving) \cite{c10}. We follow the training and test partition as \cite{c10,c14}, in which the first 800 frames are conducted as training samples with the remaining as a testing set. The crowd prediction result by our proposed approach is shown in Fig. 6, from which it can be easily observed that our method can perfectly predict the crowd number despite huge fluctuations of crowd density.

 \begin{figure*}[thpb]
      \centering
       \includegraphics[scale=0.65]{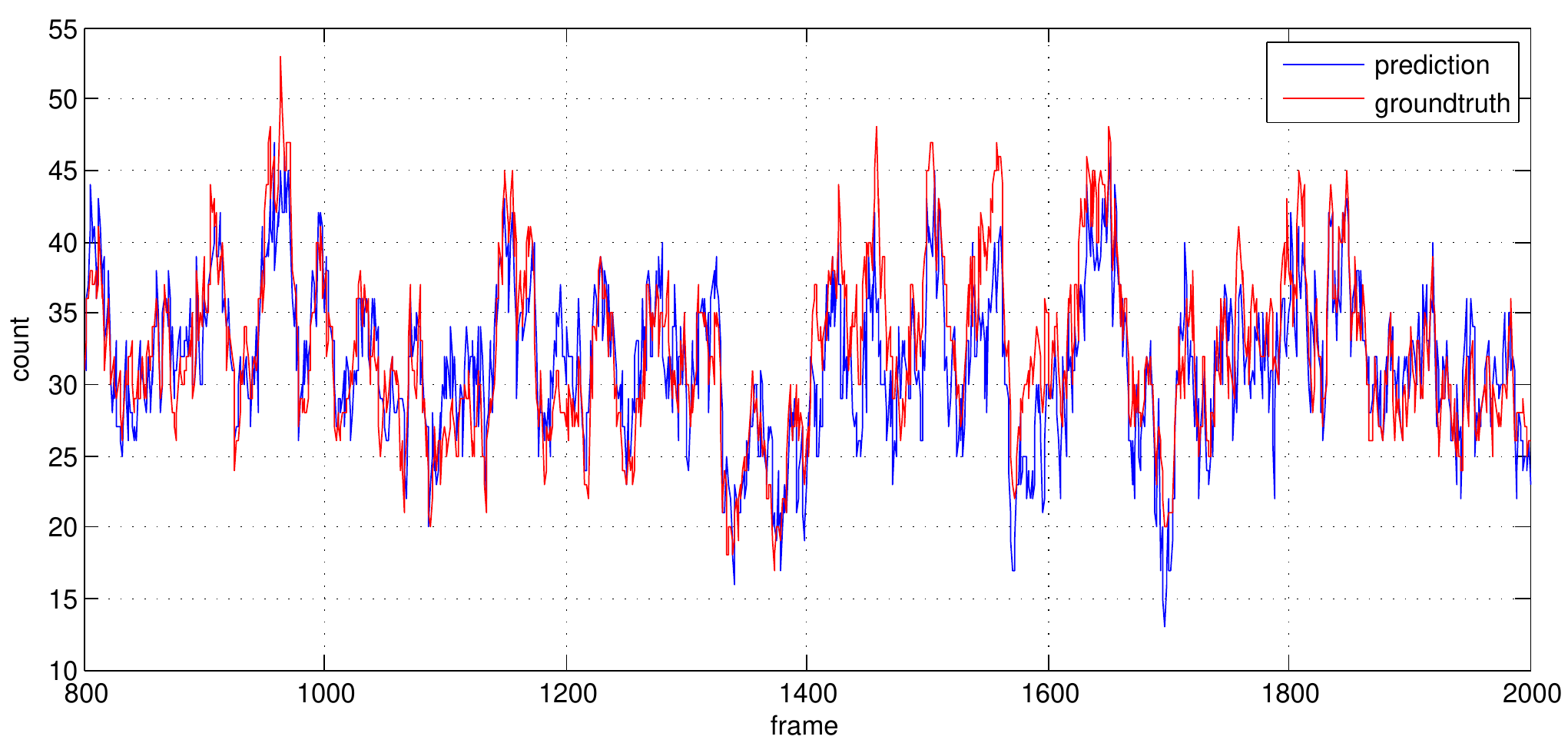}
      \caption{The prediction result for Mall dataset.}
      \label{figurelabel}
   \end{figure*}

We utilize parameters shown in the first line of Table \uppercase\expandafter{\romannumeral2} and plot the curve of MAE/MSE vs. the size of nearest neighbors $\kappa$ in Fig. 7.  Experimental results are affected by the parameter $\kappa$ selection and in the Mall dataset we set $\kappa=10$ for better performance.

\begin{figure*}[!htb]
\begin{minipage}{0.5\linewidth}
\centering
\includegraphics[scale=0.55]
{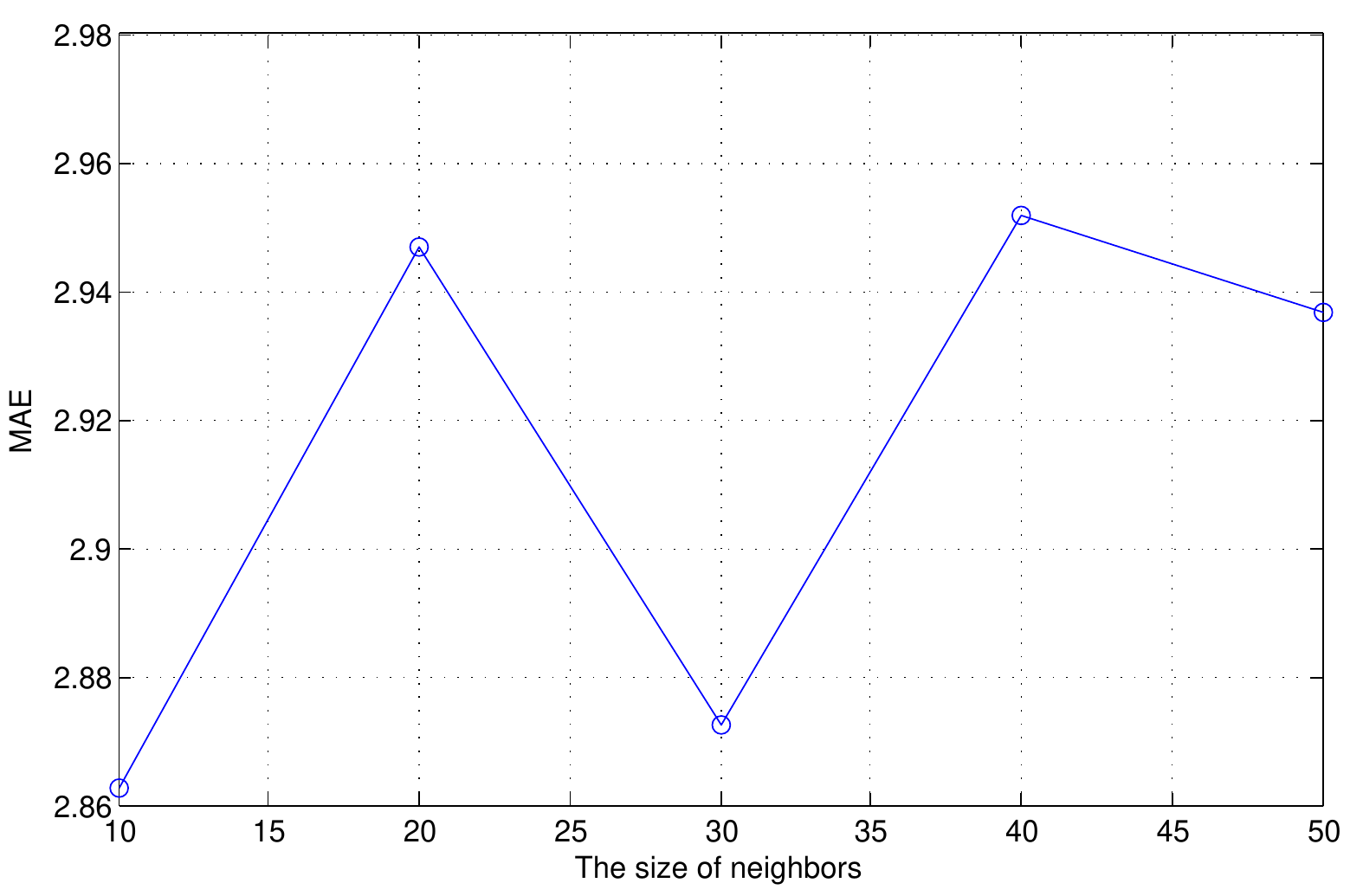}\\
\text{(a)}
\end{minipage}
\begin{minipage}{0.48\linewidth}
\centering
\includegraphics[scale=0.55]
{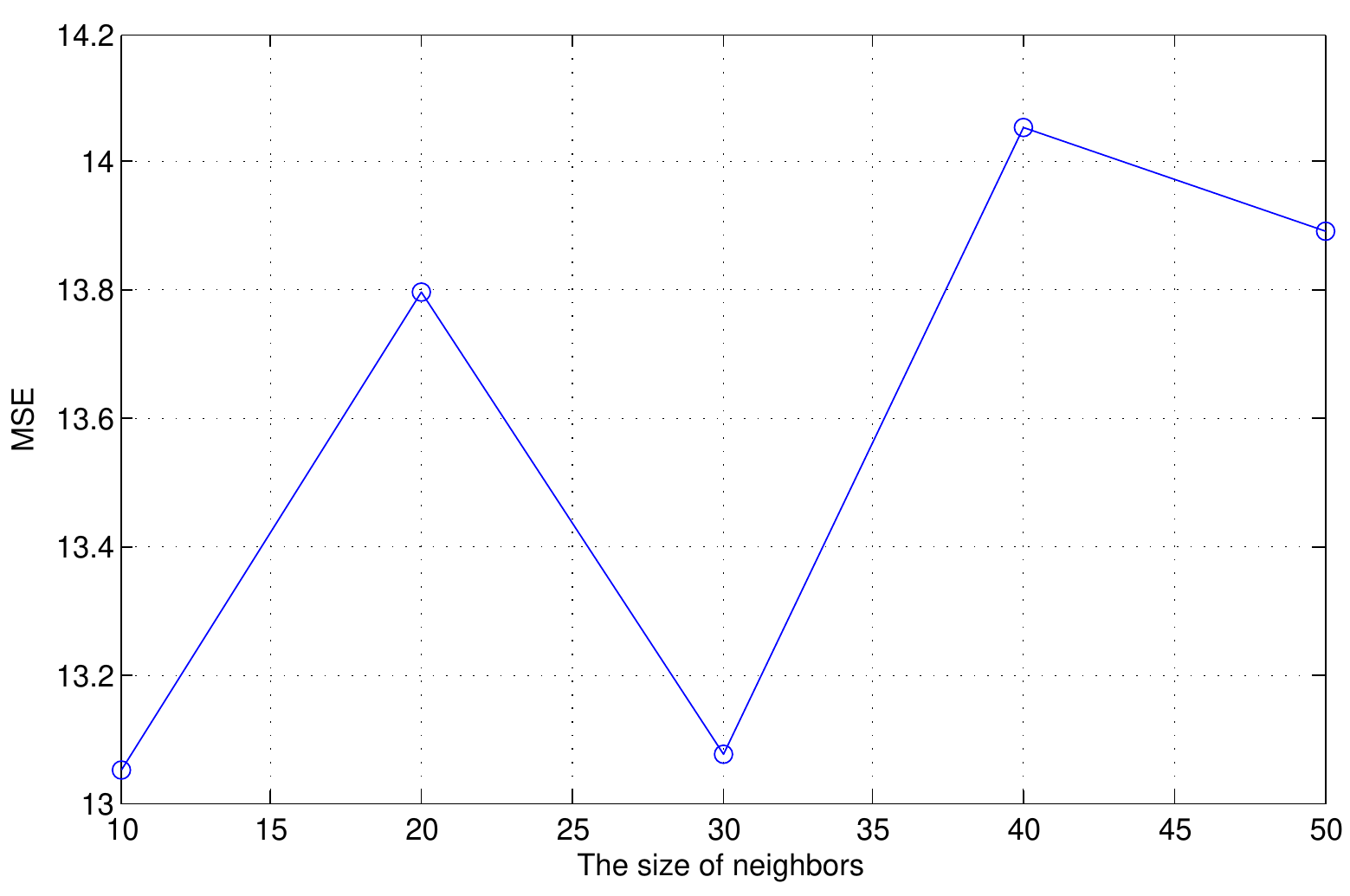}\\
\text{(b)}
\end{minipage}
\centering
  \caption{Two evaluation measures MAE \& MSE with respect to nearest neighbor sizes on the Mall dataset.}
  \label{fig8}
\end{figure*}

Table \uppercase\expandafter{\romannumeral3} shows comparison results by our approach and some existing methods. It can be easily seen from Table \uppercase\expandafter{\romannumeral3} that the proposed method outperforms all published works and achieves the state-of-the-art. It validates that the learned crowd representation is highly representative and discriminative for the Mall dataset.

\begin{table}[h]
\caption{Comparison with previous work on the Mall dataset.}
\label{table_example}
\begin{center}
\begin{tabular}{|c|c|c|}
\hline
Method & MAE  & MSE \\
\hline
MLR \cite{c15} &  3.90   & 23.9   \\
\hline
GPR \cite{c9} & 3.72  & 20.1 \\
\hline
MORR \cite{c10}  & 3.15 & 15.7 \\
\hline
NCA-RR \cite{c14}  & 4.31 & 25.8 \\
\hline
CA-RR \cite{c14}  & 3.43 & 17.7 \\
\hline
\textbf{Our method}  & \textbf{2.86} & \textbf{13.05} \\
\hline
\end{tabular}
\end{center}
\end{table}

Table \uppercase\expandafter{\romannumeral4} lists the performance of our method and baselines. The simplest representation HF has been comparable to or even exceeded some methods listed in Table \uppercase\expandafter{\romannumeral3}, which indicates the effectiveness of our pixel-level attribute feature map. Incorporating the spatial information, SPPF significantly outperforms HF using two measures. The phenomenon that SPPF is still inferior to our proposed method shows the weakness of low-level holistic features.

Further, we  combine LAF and VLAD to gain a high-level representation LFV. One evaluation index MAE falls in between HF and SPPF while MSE is superior to the above methods. The fact shows that localized methods make the distribution of errors between groundtruth and prediction be more uniform than that of holistic approaches. Next, the proposed W-VLAD is applied instead of original VLAD, the result of which is shown in the last row of Table \uppercase\expandafter{\romannumeral4} (namely our method) and reports the best performance under our framework.

\begin{table}[h]
\caption{Experimental results of baselines on the Mall dataset.}
\label{table_example}
\begin{center}
\begin{tabular}{|c|c|c|}
\hline
Method & MAE  & MSE \\
\hline
HF &  3.93  & 24.28   \\
\hline
SPPF & 3.17  & 20.84 \\
\hline
LFV  & 3.46 & 18.44 \\
\hline
\textbf{our method}  & \textbf{2.86} & \textbf{13.05} \\
\hline
\end{tabular}
\end{center}
\end{table}

\begin{figure*}[!htb]
\begin{minipage}{0.9\linewidth}
\centering
\includegraphics[scale=0.5]
{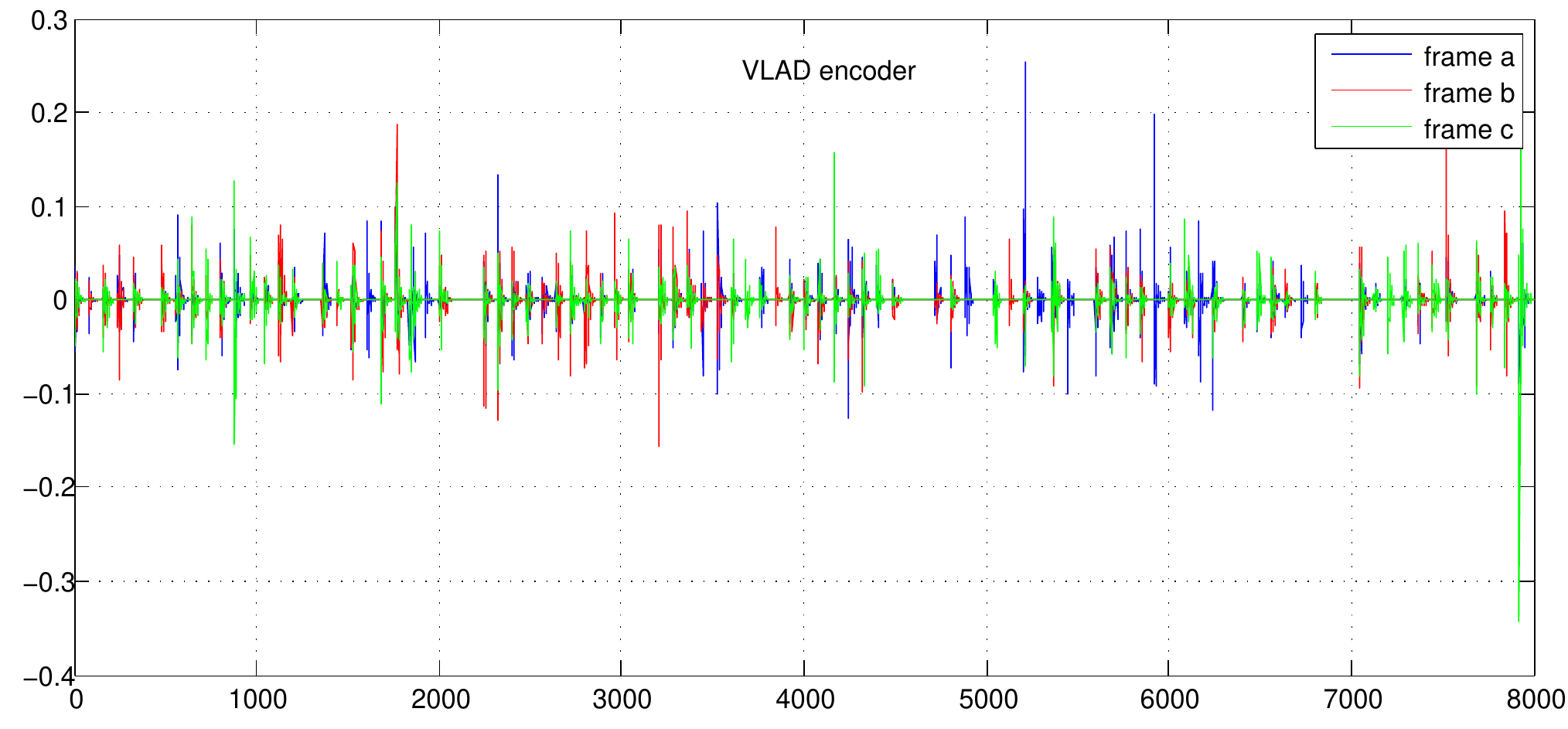}\\
\text{(a)}
\end{minipage}
\begin{minipage}{0.9\linewidth}
\centering
\includegraphics[scale=0.5]
{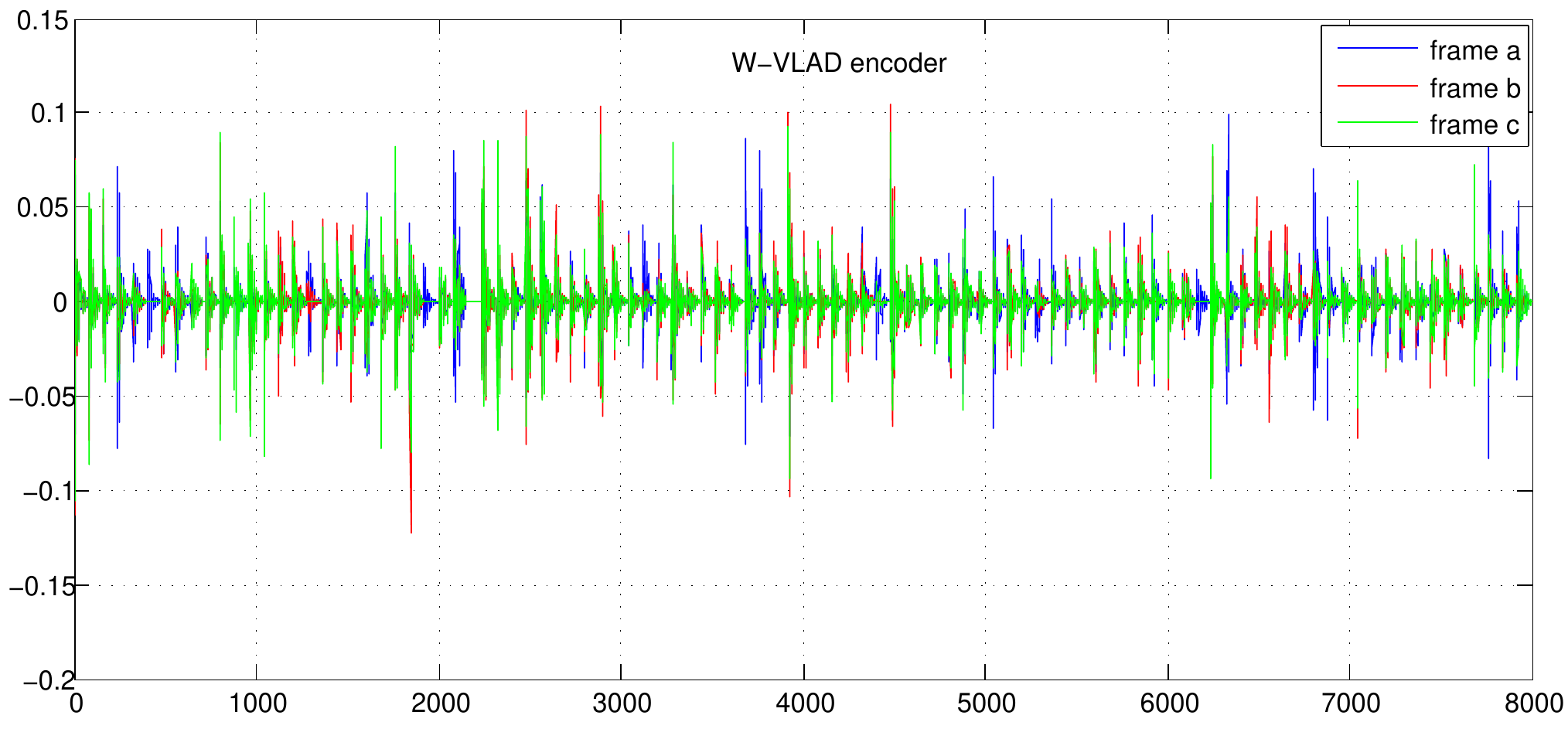}\\
\text{(b)}
\end{minipage}
\centering
  \caption{Representations obtained by VLAD and W-VLAD with the same descriptors LAF.}
  \label{fig8}
\end{figure*}

In order to show the discrimination of representations, we randomly choose three frames with different crowd numbers (frame a: 19, frame b: 30, frame c: 30) and lists corresponding similarities in Table \uppercase\expandafter{\romannumeral5}, in which the linear function is utilized here to evaluate the similarity between two vectors. The data show that the inter-class distance is larger than the inner-class distance. By our proposed method, the difference between inter-class and inner-class distance is the largest among all baselines, which validates the discriminative power of the final representation by our proposed approach.

\begin{table}[h]
\caption{The similarities between different representations.}
\label{table_example}
\begin{center}
\begin{tabular}{|c|c|c|c|c|}
\hline
   &HF   & SPPF  & LFV  & \textbf{our method} \\
\hline
S(a,b) & 0.9547  & 0.7598  & 0.0857    & 0.6261   \\
\hline
S(b,c)  & 0.9953  & 0.8985  & 0.1807  & 0.8074\\
\hline
S(a,b)-S(b,c)  & 0.0406  & 0.1387  & 0.0950 & 0.1813 \\
\hline
\end{tabular}
\end{center}
\end{table}

Taking LFV and our proposed method for example, we intuitively plot the corresponding representations shown in Fig. 8. By W-VLAD encoder, representations between frames with the same crowd number (frame b \& frame c) are more similar  and those with different crowd numbers (frame a \& frame b) have larger differences.

\begin{table}[h]
\caption{Comparison with previous work on the UCSD dataset.}
\label{table_example}
\begin{center}
\begin{tabular}{|c|c|c|}
\hline
Method & MAE  & MSE \\
\hline
MLR \cite{c15} &  2.60   & 10.1   \\
\hline
\textbf{GPR \cite{c9}}  & \textbf{2.24}  &\textbf{ 7.97 }\\
\hline
MORR \cite{c10}  & 2.29 & 8.08 \\
\hline
NCA-RR \cite{c14}  & 2.85 & 11.9 \\
\hline
RFR \cite{c14}   & 2.42& 8.47 \\
\hline
Our method  & 2.41 & 9.12 \\
\hline
\end{tabular}
\end{center}
\end{table}

\subsection{UCSD dataset}
The UCSD dataset is recorded at a campus scene by an  hand-held camera. It contains 2000 frames, 610-1400 of which are employed as a training set and the rest for testing. Table \uppercase\expandafter{\romannumeral6} lists results by previous methods and our approach. It can be seen that MAE/MSE achieves 2.41/9.12, which is a comparable result.

\begin{table}[h]
\caption{Experimental results of baselines on the UCSD dataset.}
\label{table_example}
\begin{center}
\begin{tabular}{|c|c|c|}
\hline
Method & MAE  & MSE \\
\hline
HF &  3.51  & 18.70   \\
\hline
SPPF & 3.47  & 17.46 \\
\hline
LFV  & 3.37 & 18.14 \\
\hline
\textbf{our method}  & \textbf{2.41} & \textbf{9.12} \\
\hline
\end{tabular}
\end{center}
\end{table}

Baselines are also utilized to verify effectiveness of our proposed parts. As shown in in Table \uppercase\expandafter{\romannumeral7}, SPPF achieves a more superior result than HF since spatial cue is taken into consideration. Compared with SPPF, LFV can obtain improved MAE and comparable MSE because constructing final representation by encoding local descriptors can retain more locality information than  holistic representation. For the UCSD dataset, results of our method and LFV show that computing weights by soft-assignment coding in W-VLAD is more effective than binary assignments in the original VLAD.

\subsection{Caltech 10X dataset}
Although the crowd density is more sparse than Mall and UCSD, it is more difficult to pick out pedestrians because of severe occlusion, the changing scenes and so on. As shown in the bottom row of Fig. 5, sometimes we cannot even identify persons from images by our naked eyes. We choose the first 2000 frames that contain more than six persons in the Caltech training dataset as the entire crowd counting dataset, in which the first 800 images are used for training with others for testing. The annotations provided in Caltech are directly utilized as crowd groundtruth.

\begin{figure}[!htb]
\begin{minipage}{0.5\linewidth}
\centering
\includegraphics[width=4cm,height=3cm]
{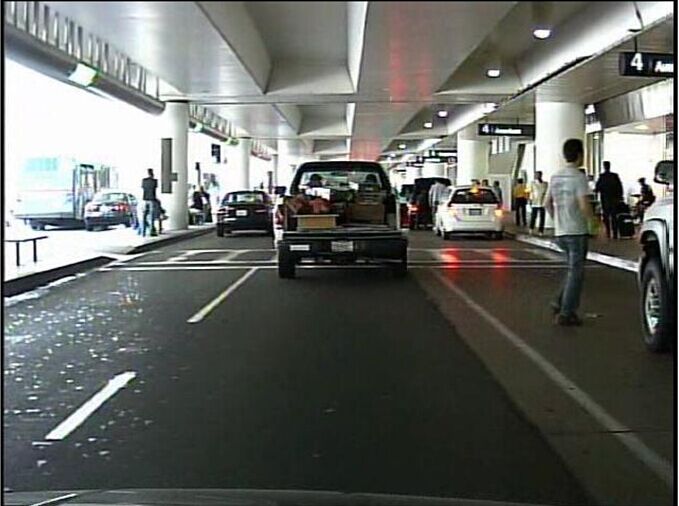}\\
\text{(a)}
\end{minipage}
\begin{minipage}{0.47\linewidth}
\centering
\includegraphics[width=4cm,height=3cm]
{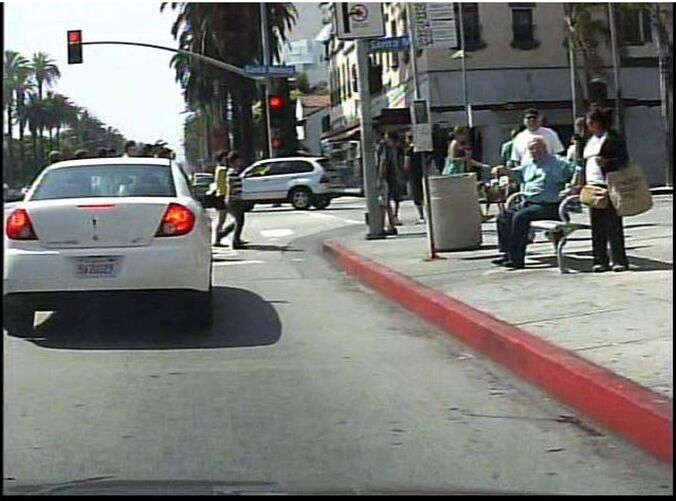}\\
\text{(b)}
\end{minipage}
\centering
  \caption{Two inaccurately predicted examples in the Caltech dataset.}
  \label{fig8}
\end{figure}

Parameters used for the Caltech dataset are shown in Table \uppercase\expandafter{\romannumeral2}. The experimental result (mae \& mse: 1.36 \& 3.50) demonstrates that our method can predict the crowd number with minor errors in most frames of Caltech dataset. Fig. 9 shows two inaccurately predicted examples that exhibit inferior performance. It can be observed that the background, such as a pole, is easily mistaken for a person in dark scenes. Besides, there exists severe dynamic changes (e.g, from the parking lot to the road at the traffic lights) in the background of crowd scene, which also makes it difficult to estimate the number of people accurately.

\begin{table}[h]
\caption{Experimental results on the Caltech dataset.}
\label{table_example}
\begin{center}
\begin{tabular}{|c|c|c|}
\hline
Method & MAE  & MSE \\
\hline
LBP &  1.62  & 4.51   \\
\hline
HF & 1.45  & 4.14 \\
\hline
LFV  & 1.40 & 3.54 \\
\hline
\textbf{our method}  & \textbf{1.36} & \textbf{3.50} \\
\hline
\end{tabular}
\end{center}
\end{table}

Quantitative results of crowd counting on Caltech dataset  and that of our method are reported in Table \uppercase\expandafter{\romannumeral8}. First, we extract LBP features to predict the crowd number, the corresponding results of which are shown in the top row.  Our method only beats LFV by a minor margin. The possible reason is that for the complicated Caltch dataset different weight assignments may be not as important as those for easier datasets such as Mall and UCSD. The comparison results shown in Table \uppercase\expandafter{\romannumeral8} verify the superiority of our learned representation.

\section{Conclusions}
In this paper, we introduce a novel method to obtain representations for crowd frames. This is the first work that attempts to utilize pixel-wise attribute feature map to describe a crowd scene. Completely different from previous features, locality-aware features with spatial cues (LAF) are extracted from the dense attribute feature map, which has better capability for describing semantic context information in crowd scenes. We further apply an improved version of VLAD, namely W-VLAD, which considers different coefficient weights for  deviations between descriptors and clusters. The discriminative power and effectiveness of our proposed method are validated on three benchmarks.

\section*{Acknowledgements}

This work is supported by National Natural Science Foundation (NNSF) of China under Grant. 61473086, 61375001 partly supported by the open fund of Key Laboratory of Measurement and partly supported by Control of Complex Systems of Engineering, Ministry of Education (No. MCCSE2013B01), the NSF of Jiangsu Province (Grants No BK20140566, BK20150470) and China Postdoctoral science Foundation
(2014M561586).

\textit{}

\end{document}